\documentclass[a4paper]{llncs}

\usepackage{amssymb}
\usepackage{graphicx}
\usepackage{amsmath}
\usepackage{float}
\usepackage{color}

\begin{document}

\mainmatter  

\title{Simultaneous Multiple Surface Segmentation Using Deep Learning\\}

\author{Abhay Shah$^{1}$, Michael Abramoff$^{1,2}$
\and Xiaodong Wu$^{1,3}$}

\institute{Department of $^{1}$Electrical and Computer Engineering, $^{2}$Radiation
Oncology, $^{3}$Department of Ophthalmology and Visual Sciences, University of Iowa, Iowa City, USA
\url{}}

\maketitle

\begin{abstract}
The task of automatically segmenting 3-D surfaces representing
boundaries of objects is important for quantitative analysis of volumetric images,
and plays a vital role in biomedical image analysis. 
Recently, graph-based methods with a global optimization property have
been developed and optimized for various medical imaging applications. 
Despite their widespread use, these require
human experts to design transformations, image features, surface smoothness priors,
and re-design for a different tissue, organ or imaging modality. 
Here, we propose a Deep Learning based approach
for segmentation of the surfaces in volumetric medical images, by learning the
essential features and transformations from training data, without any human expert intervention.
We employ a regional approach to learn the local surface profiles.
The proposed approach was evaluated on simultaneous intraretinal layer segmentation of optical
coherence tomography (OCT) images of normal retinas and retinas affected by age related macular
degeneration (AMD). The proposed approach was validated on 40 retina OCT volumes 
including 20 normal and 20 AMD subjects.
The experiments showed statistically significant improvement
in accuracy for our approach compared to state-of-the-art
graph based optimal surface segmentation with convex priors (G-OSC).
A single Convolution Neural Network (CNN) was used to learn the surfaces for both normal
and diseased images.
The mean unsigned surface positioning errors obtained by 
G-OSC method 
$2.31$ voxels ($95\%$ CI $2.02$-$2.60$ voxels) was
 improved to $1.27$ voxels ($95\%$ CI $1.14$-$1.40$ voxels)
using our new approach.
On average, our approach takes $94.34$ s, requiring 95.35 MB memory,
 which is much faster than the $2837.46$ s and 6.87 GB memory
 required by the G-OSC method on the same computer system. 
 
\keywords{Optical Coherence Tomography (OCT), deep learning,
 Convolution Neural Networks (CNNs), multiple surface segmentation}
 
\end{abstract}

\section{Introduction}
\begin{figure} 
\centering
\includegraphics[width=8cm]{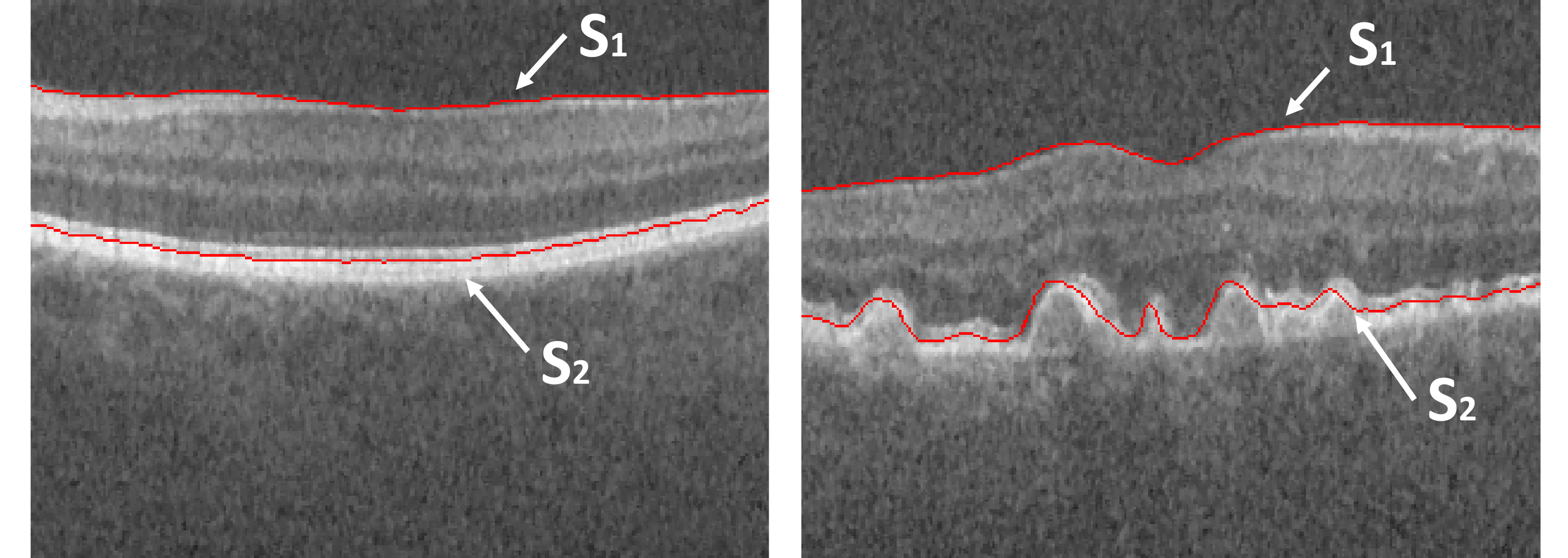}
\caption{Illustration of difference in surface profiles on a single B-scan. (left) Normal Eye (right) Eye with AMD.
 $S_{1}$ = ILM and $S_{2}$ = IRPE, are shown in red.}
\label{fig:1}
\end{figure}

For the diagnosis and management of disease, segmentation of
 images of organs and tissues is a crucial step
 for the quantification of medical images.
 Segmentation finds the boundaries or, limited to the
 3-D case, the surfaces, that separate organs, tissues
 or regions of interest in an image.
Current state-of-the-art methods for automated 3-D surface segmentation use expert
 designed graph search / graph cut approaches \cite{lee2010automated} or
 active shape/contour modelling \cite{yazdanpanah2009intra}, all based on classical
 expert designed image properties,
 using carefully designed transformations
 including mathematical morphology and wavelet transformations.
 For instance, OCT is a 3-D imaging technique that is widely used in the diagnosis and management
 of patients with retinal diseases. The tissue boundaries in OCTs vary by 
 presence and severity of disease. An example is shown in Fig.\ref{fig:1}(a)(b) to
 illustrate the difference in profile for the Internal Limiting Membrane (ILM) and Inner 
Retinal Pigment Epithelium (IRPE) in a normal eye and in an eye with AMD. 
 In order to overcome these different manifesttaions, graph based
 methods \cite{lee2010automated}\cite{shah2015multiple}\cite{song2013} with
 transformations,
 smoothness constraints, region of interest extraction and multiple resolution approaches
 designed by experts specifically
 for the target surface profile have been used. The 
 contour modelling approach \cite{yazdanpanah2009intra} requires
 surface specific and expert designed region based and shape prior terms. 
 The current methods for surface segmentation in OCTs
are  highly dependent on expert designed target
 surface specific transformations and therefore, there is a desire for approaches which do not require human expert intervention. 
Deep learning, where all transformation levels
 are determined from training data, instead of being
 designed by experts, has been
 highly successful in a large number of computer vision \cite{krizhevsky2012imagenet}
 and medical image
 analysis detection tasks \cite{brats}\cite{kaggle}, substantially outperforming all classical
 image analysis techniques, and given
 the spatial coherence that is characteristic of images, typically
 implemented as Convolutional Neural Networks (CNN)
 \cite{krizhevsky2012imagenet}.
All these
 examples are where CNNs are used to identify pixels or voxels
 as belonging to a certain class, called classification
 and not to identify boundaries in the
 images, i.e. segmentation.
 
  In this study, we propose a CNN based deep learning approach for
  boundary surface segmentation of a target object,
 where  both features and models are learnt from training data
 without intervention by a human expert. 
 We are particularly interested in terrain-like surfaces.
 An image volume is generally represented as a 3-D volumetric
 cube consisting of voxel columns, wherein a terrain-like surface intersects each column
 at exactly one single voxel location. 
 The smoothness of a given surface may be interpreted as the piecewise change in the
 surface positions of two
neighboring columns. The graph based optimal surface segmentation
 methods \cite{lee2010automated}\cite{song2013}
use convex functions
to impose the piecewise smoothness while globally minimizing the objective function for segmentation.
In order to employ CNNs for surface segmentation,
 two key questions need to be answered. First, since most of the
 CNN based methods have been used for classification,
 how can a boundary be segmented using a CNN? Second, how can the CNN learn
 the surface smoothness and surface distance between two interacting surfaces implicitly?
 We answer these questions by representing
 consecutive target surface positions for a given input
 image as a vector. For example, $m_{1}$ consecutive target
 surface positions for a single surface are represented as a $m_{1}$-D vector, which may be
 interpreted as a point in the $m_{1}$-D space, while maintaining a strict
 order with respect to the consecutiveness of the target surface
 positions. The ordering of the target surface positions partially
 encapsulates the smoothness of the surface.  
 Thereafter, the error
 (loss) function utilized in the CNN to back propagate the error is
 chosen as a Euclidean loss function shown in Equation~(\ref{eqn:2}),
 wherein the network adjusts the weights of the various nonlinear
 transformations within the network to minimize the Euclidean distance
 between the CNN output and the target surface positions in the $m_{1}$-D
 space. 
 Similarly, for detecting $\lambda$ surfaces, the surface positions are represented
 as a $m_{2}$-D vector, where $\lambda = \{1, 2, \ldots
\lambda\}$, $m_{2} = \lambda \times m_{1}$ and $m_{1}$ consecutive
 surface poistions for a
surface index $i$ ($ i \in \lambda$)
are given by $\{((i-1) \times m_{1})+1, ((i-1) \times m_{1})+2, \ldots ((i-1) \times m_{1})+m_{1}\}$
 index elements in the $m_{2}$-D vector.
 
In the currently used methods for segmentation of surfaces,
 the surface smoothness is piecewise in
 nature. The surface smoothness penalty (cost) enforced
 in these methods is the sum of the surface smoothness penalty
 ascertained using the difference of two consecutive surface positions.
 Thus, such methods require an expert designed, and application specific,
 smoothness term to attain
 accurate segmentations. 
On the contrary, segmentation using CNN should be expected to also
 learn the different smoothness profiles of the target surface. Because
 the smoothness is piecewise, it should be sufficient for the CNN to
 learn the different local surface profiles for individual segments of
 the surface with high accuracy because the resultant surface is a combination of these
 segments.
 Hence, the CNN is trained on individual
 patches of the image with segments of the target surface. 

 

 \section{Method}\label{sec:method} 
\begin{figure} 
\centering
\includegraphics[width=8cm, height = 3cm]{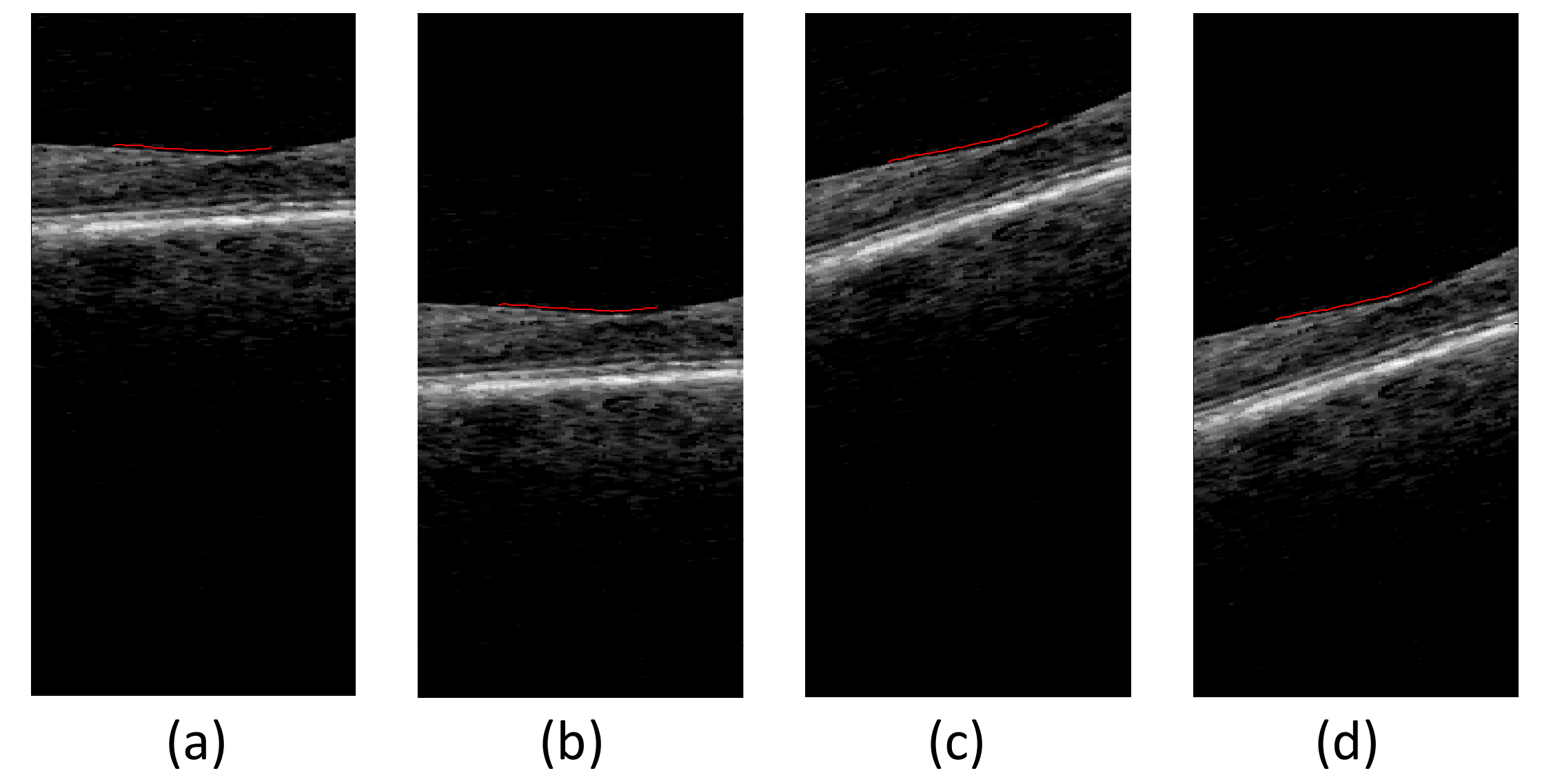}
\caption{Illustration of data augmentation applied to an input patch. 
The target surface is shown in red. (a)Extracted patch from a B-scan, (b)Translation (c)Rotation (d)Translation and rotation, as applied to (a).}
\label{fig:3}
\end{figure}

 Consider a volumetric image $I(x,y,z)$ of size $X\times Y\times Z$.
A surface is defined as $S(x,y)$, where $x \in {\bf x}$ = $\{0,1,...X-1\}$, 
$y \in {\bf y}$ =$\{0,1,...Y-1\}$ and $S(x,y) \in {\bf z}$ = $\{0,1,...Z-1\}$.
Each $(x,y)$ pair forms a voxel column parallel to the z-axis, wherein the surface $S(x,y)$ intersects each column
 at a single voxel location. For simlutaneously segmenting $\lambda (\lambda \geq 2)$ surfaces, the 
goal of the CNN is to learn the surface positions $S_{i}(x,y)$ ($i \in \lambda$) for columns formed by each $(x,y)$ pair.
In this work, we present a slice by slice segmentation of a 3-D
 volumetric image applied on OCT volumes. 
 Patches are extracted from B-scans with the target
 Reference Standard (RS).
 A patch $P(x_{1},z)$ is of size $N\times Z$, where $x_{1} \in {\bf x_{1}}$ = $\{0,1,...N-1\}$,
 $z \in {\bf z}$ =$\{0,1,...Z-1\}$
 and $N$ is a multiple of $4$. The target surfaces $S_{i}$'s
 to be learnt simultaneously from $P$ is $\overline{S_{i}}(x_{2}) \in {\bf z}$ = $\{0,1,...Z-1\}$, where
 $x_{2} \in {\bf x_{2}}$ = $\{\frac{N}{4},\frac{N}{4}+1...\frac{3N}{4}-1\}$. Essentially,
 the target surfaces to be learnt is the surface locations for the middle $\frac{N}{2}$ consecutive
 columns in $P$. The overlap between consecutive patches 
 ensures no abrupt changes occur at patch boundaries. 
 By segmenting the middle $N/2$ columns in a patch size with $N$ columns,
 the boundary of patches overlap with the consecutive surface segment patch. 
Then, data augmentation is
 performed, where
 for each
 training patch, three additional training patches were created. 
 First, a random translation value was chosen between -250 and 250 such that the
 translation was within the range of the patch size. The training patch and the
 corresponding RS for surfaces $\overline{S_{i}}$'s were translated in the $z$ dimension accordingly.
 Second, a random rotation value was chosen between -45 degrees and 45 degrees. The training patch
 and the corresponding RS for surfaces $\overline{S_{i}}$'s were rotated accordingly.
 Last, a combination of rotation and translation was used to generate another patch.
 Examples of data augmentation on patches for a single surface is shown in Fig.\ref{fig:3}.



\begin{figure} 
\centering
\includegraphics[width=11cm]{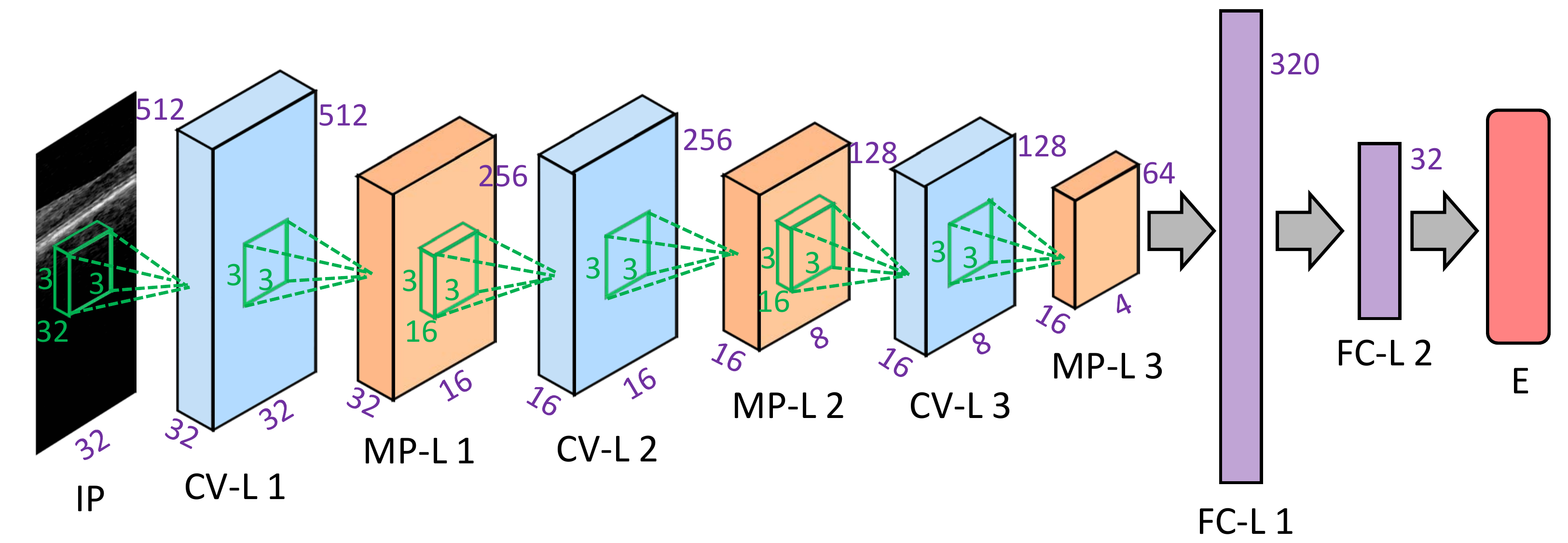}
\caption{The architecture of the CNN learned in our work for $N$=$32$ and $\lambda =2$.
 The numbers along each side of the cuboid indicate the dimensions of the feature maps.
 The inside cuboid ($green$) represents the convolution kernel
 and the inside square ($green$) represents the pooling region size.
 The number of hidden neurons in the fully connected layers are marked aside.
 IP=Input Patch, CV-L=Convolution Layer, MP-L=Max-Pooling Layer, FC-L=Fully Connected Layer,
 E=Euclidean Loss Layer.}
\label{fig:4}
\end{figure}

For segmenting $\lambda$ surfaces simultaneously, the CNN learns
$\lambda$ surfaces for each patch.
The CNN architecture used in our work is shown in Fig.\ref{fig:4}, employed for $\lambda =2$ and 
patches with $N=32$. 
The CNN contains three convolution layers \cite{krizhevsky2012imagenet}, 
each of which is followed by a max-pooling layer \cite{krizhevsky2012imagenet} with
stride length of two.
Thereafter, it is followed by two fully connected layers \cite{krizhevsky2012imagenet}, where
the last fully connected layer represents the final output of the middle $\frac{N}{2}$ surface
 positions for 2 target surfaces in $P$. 
 Lastly, a Euclidean loss function (used for regressing to real-valued labels)
 as shown in Equation~(\ref{eqn:2}) is
 utilized to compute the error between CNN outputs
 and RS of $S_{i}$'s ($i \in \lambda$) within $P$ for back propagation during the training phase.
Unsigned mean surface positioning error (UMSPE) \cite{lee2010automated} is one of the commonly used
 error metric for evaluation of surface segmentation accuracy.
 The Euclidean loss function (E), essentially computes sum
 of the squared unsigned surface positioning error over the $\frac{N}{2}$ consecutive
 surface position for $S_{i}$'s of the CNN output and the RS for $P$.

\begin{equation}
E =  \sum_{i =1}^{i =\lambda} \sum_{k_{1}=0}^{k_{1} = \frac{N}{2} -1} (\overline{a}^{i}_{k_{1}} - a^{i}_{k_{2}})^{2}
\label{eqn:2}
\end{equation}
where $k_{2} = ((i-1) \times N/2) + k_{1}$, $\overline{a}^{i}_{k_{1}}$ and ${a}^{i}_{k_{2}}$ is the $k_{1}$-$th$ surface position of reference standard and CNN output respectively for 
surface $S_{i}$ in a given $P$.

\section{Experiments}
The experiments compare
segmentation accuracy of the proposed CNN based method (CNN-S) and
G-OSC method \cite{song2013}.
The two surfaces simultaneously segmented in this
study are $S_{1}$-ILM and $S_{2}$-IRPE
as shown in Fig. \ref{fig:1}.
$115$ OCT scans of normal eyes, $269$ OCT scans of eyes with
AMD and their respective reference standards (RS) (created by experts 
with aid of the DOCTRAP software \cite{farsiu2014quantitative})
were obtained from the publicly available repository 
\cite{farsiu2014quantitative}.
The 3-D
volume size was $1000 \times 100 \times 512$ voxels.
The data volumes were divided into a training set ($79$ normal and $187$ AMD),
a testing set ($16$ normal and $62$ AMD)
and a validation set ($20$ normal and $20$ AMD).
The volumes were denoised by applying a median filter of
size $5 \times 5 \times 5$ and normalized
with the resultant voxel intensity varying from -1 to 1.
Thereafter, patches of size $N \times 512$ with
their respective RS
for the middle $\frac{N}{2}$ consecutive surface
positions for $S_{1}$ and $S_{2}$ is extracted using data augmentation, for
training and testing volumes, resulting in a
training set of $340,000$ and testing set of $70,000$ patches.
In our work, we use $N=32$.
The UMSPE
was used to evaluate 
the accuracy. The complete surfaces for each validation volume
 were segmented using the CNN-S method by creating $\frac{1016}{N/2}$ patches from each B-scan where
 each B-scan was zero padded with $8$ voxel columns at each extremity. 
 Statistical significance
of observed differences was determined by paired Student $t$-tests for
which $p$ value of 0.05 was considered significant.
In our study we used one NVIDIA Titan X GPU for training the CNN.
The validation using the G-OSC and CNN-S method were carried out on a
on a Linux workstation (3.4 GHz, 16 GB memory).
A single CNN was trained to infer on both the normal and AMD OCT scans.
For a comprehensive comparison, three experiments were performed with the G-OSC method. The first 
experiment (G-OSC $1$) involved segmenting the surfaces in both normal and AMD OCT scans using a
same set of optimized parameters. The second (G-OSC $2$) and third (G-OSC $3$) experiment
involved segmenting the normal and AMD OCT scans with different set of optimized parameters, respectively.

\section{Results}
\begin{figure} 
\centering
\includegraphics[width=12.5cm]{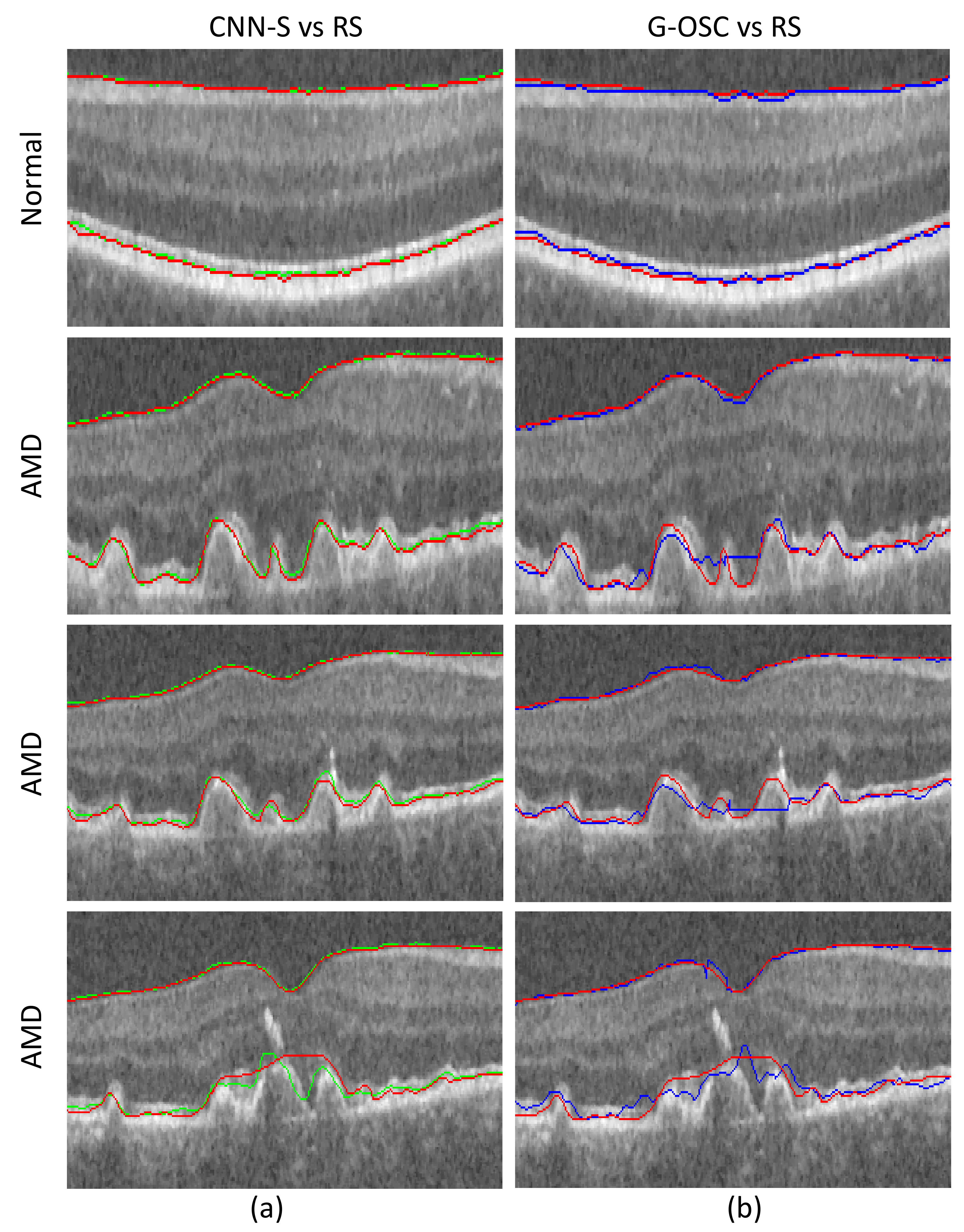}
\caption{Each row shows the same B-scan from a Normal or AMD OCT volume. (a) CNN-S vs. RS 
(b) G-OSC vs. RS, for surfaces $S_{1}$ =ILM and $S_{2}$ = IRPE. RS = Reference Standard, Red = reference standard, Green = 
Segmentation using proposed method and Blue = Segmentation using G-OSC method. 
In the 4th row, we had the reference standard reviewed by a fellow-ship trained retinal specialist,
who stated that the CNN-S method is closer to the real surface than the reference standard.}
\label{fig:5}
\end{figure}

 The quantitative comparisons between the proposed CNN-S method
and the G-OSC method 
 on the validation volumes is summarized in Table~\ref{table:1}. 
 For the entire validation data, the proposed method produced significantly lower UMSPE
for surfaces $S_{1}$ ($p<0.01$) and $S_{2}$ ($p<0.01$), compared to the segmentation 
results of G-OSC $1$, G-OSC $2$ and G-OSC $3$.
Illustrative results of segmentations from
the CNN-S, G-OSC $2$ and G-OSC $3$ methods on validation volumes
are shown in Fig.~\ref{fig:5}.
It can be observed that CNN-S method yeilds consistent and qualitatively
superior segmentations with respect to the G-OSC method.
On closer analysis of some B-scans in the validation data,
the CNN-S method produced high quality segmentation for 
a few cases
where the RS was not
accurate enough as verifed by an expert (4th row in Fig.~\ref{fig:5}). 
The CNN required $17$ days to train on the GPU.
The CNN-S method with average computation time of 94.34 seconds (95.35 MB memory) is
much faster than G-OSC with average computation time of 2837.46 seconds (6.87 GB memory). 
 
\begin{table}[ht]
\scriptsize
 \caption{UMSPE expressed as (mean $\pm$ $95\%$ CI) in voxels. RS - Reference Standard.
 $N = 32$ was used as the patch size ($32 \times 512$).}
\centering
\begin{tabular}{|c|c|c||c|c|c|c|}
\hline
 & \multicolumn{2}{|c||}{Normal and AMD} & \multicolumn{2}{|c|}{Normal} & \multicolumn{2}{|c|}{AMD}\\
\hline 
 & \ \  G-OSC 1  \ & \ \  CNN-S \ & \ \  G-OSC 2 \ \ \ & CNN-S \ \ & \ G-OSC 3 \ \ \ & CNN-S \\
  Surface & \ \  vs. RS \ & \ \ vs. RS \ & \ \ vs. RS \ \ \ & vs. RS \ \ & \ vs. RS\ \ \ & vs.RS\\[0.5ex]   
\hline
$S_{1}$ & 1.45 $\pm$ 0.19 & 0.98 $\pm$ 0.08 & 1.19 $\pm$ 0.05 & 0.89 $\pm$ 0.07 & 1.37 $\pm$ 0.22 & 1.06 $\pm$ 0.11\\
$S_{2}$ & 3.17 $\pm$ 0.43 & 1.56 $\pm$ 0.15 & 1.41 $\pm$ 0.11 & 1.28 $\pm$ 0.10 & 2.88 $\pm$ 0.54 & 1.83 $\pm$ 0.26\\
\hline
Overall & 2.31 $\pm$ 0.29 & 1.27 $\pm$ 0.13 & 1.31 $\pm$ 0.07 & 1.08 $\pm$ 0.08 & 2.13 $\pm$ 0.39 & 1.44 $\pm$ 0.19\\
\hline
\end{tabular}
\label{table:1}
\end{table}
\normalsize

\section{Discussion and Conclusion}
The results demonstrate 
superior quality of segmentations compared to the G-OSC method, while 
eliminating
the requirement of expert designed transforms. The proposed method used a single CNN
to learn various local surface profiles for both normal and AMD data.
Our results compared to G-OSC $1$ show that the CNN-S methods outperforms the G-OSC method.
If the parameters are tuned specifically for each type of data by using expert prior knowledge
while using the G-OSC method,
as in the cases of G-OSC $2$ and G-OSC $3$,
the results depict that the CNN-S method still results in superior performance. 
The inference using CNN-S is much faster than G-OSC method and requires much less memory.
Consequently, the inference can be parallelized for multiple patches, thereby
further reducing the computation
time, thus making it potentially more suitable for clinical applications.
However, a drawback of any such learning approach
in medical imaging is the limited amount of available training data.
The proposed method was trained on images from one type of scanner and hence it is possible
that the trained CNN may not produce consistent segmetnations on images
obtained from a different scanner due
to difference in textural and spatial information.
The approach can readily be extended to perform
3-D segmentations by employing 3-D convolutions.

In this paper, we proposed a CNN based method for segmentation of surfaces in
volumetric images with implicitly learned surface smoothness and surface separation models. 
We demonstrated
the performance and potential of the proposed method through application on
OCT volumes to segment the ILM and IRPE surface. The experiment results show higher
segmentation accuracy as compared to the G-OSC method. 
To the best of our knowledge, this is the first method
of its kind that does not require any human intervention for surface segmentation.  

\bibliographystyle{splncs03}
\bibliography{refrences_library}

\end{document}